\documentclass[sigconf]{acmart}

\AtBeginDocument{%
  }

\usepackage{graphicx}

\copyrightyear{2024}
\acmYear{2024}
\setcopyright{rightsretained}
\acmConference[SIGCSE Virtual 2024]{Proceedings of the 2024 ACM Virtual Global Computing Education Conference V. 2}{December 5--8, 2024}{Virtual Event, NC, USA}
\acmBooktitle{Proceedings of the 2024 ACM Virtual Global Computing Education Conference V. 2 (SIGCSE Virtual 2024), December 5--8, 2024, Virtual Event, NC, USA}
\acmDOI{10.1145/3649409.3691083}
\acmISBN{979-8-4007-0604-2/24/12}

\usepackage[english, status=final, margin=false, inline=true]{fixme}
\usepackage{xcolor}
\fxusetheme{color}

\FXRegisterAuthor{jf}{ajf}{\color{red}JF}

\FXRegisterAuthor{gm}{agm}{\color{blue}GM}

\newcommand{\rmspace}{\vspace{-3.8ex}}

\usepackage{pifont}

\begin{document}

\title{ASAG2024: A Combined Benchmark for Short Answer Grading}

\author{Gérôme Meyer}
\affiliation{%
  \institution{Zurich University of Applied Sciences}
  \city{Winterthur}
  \country{Switzerland}}
\email{gerome.meyer@protonmail.com}

\author{Philip Breuer}
\affiliation{%
  \institution{Zurich University of Applied Sciences}
  \city{Winterthur}
  \country{Switzerland}}
\email{philip.breuer@protonmail.com}

\author{Jonathan Fürst}
\affiliation{%
  \institution{Zurich University of Applied Sciences}
  \city{Winterthur}
  \country{Switzerland}}
\email{jonathan.fuerst@zhaw.ch}


\begin{abstract}
Open-ended questions test a more thorough understanding than closed-ended questions and are often a preferred assessment method.
However, open-ended questions are tedious to grade and subject to personal bias. Therefore, there have been efforts to speed up the grading process through automation. Short Answer Grading (SAG) systems aim to automatically score students’ answers. 
Despite growth in SAG methods and capabilities, there exists no comprehensive short-answer grading benchmark across different subjects, grading scales, and distributions. Thus, it is hard to assess the capabilities of current automated grading methods in terms of their generalizability.
In this preliminary work, we introduce the combined ASAG2024 benchmark to facilitate the comparison of automated grading systems. Combining seven commonly used short-answer grading datasets in a common structure and grading scale. For our benchmark, we evaluate a set of recent SAG methods, revealing that while LLM-based approaches reach new high scores, they still are far from reaching human performance.
This opens up avenues for future research on human-machine SAG systems.
\end{abstract}

\ccsdesc[500]{Applied computing~Education}
\ccsdesc[500]{Computing methodologies~Artificial intelligence}

\keywords{Automated Grading, ASAG, Education, Benchmark, Dataset, LLMs}


\maketitle

\section{Introduction}

Written examinations are still widely used to assess students' know-how of a subject's learning objectives. Among the various question types, open-ended questions can test a more thorough understanding compared to closed-ended questions such as multiple-choice~\cite{ozuru2013comparing}. However, grading the answers to such questions is demanding as it requires extensive manual grading effort and can be subject to personal biases. Therefore, there have been several efforts to speed up the grading process through automation. Short Answer Grading (SAG) systems aim to automatically score students’ answers in examinations.
While earlier SAG systems focused on concept mapping, information extraction, and corpus-based methods~\cite{burrows2015eras}, current systems employ fine-tuned language models~\cite{kim2024prometheus, BART-SAF} or even directly in-context learning with powerful task-independent Large Language Models (LLMs)~\cite{bubeck2023sparks}.

Despite this growth in methods and capabilities, to the best of our knowledge, \textit{there exists no comprehensive short-answer grading benchmark across different subjects, grading scales, and distributions}. Thus, assessing the generalizability of current automated grading methods is challenging.
In this preliminary work, we aim to raise awareness of this lack of a comprehensive benchmark by providing the first version of such a benchmark together with an initial evaluation of existing automated grading solutions. 

Specifically, we introduce the combined ASAG2024 benchmark\footnote{Available online: \url{https://huggingface.co/datasets/Meyerger/ASAG2024}}  to facilitate the comparison of automated grading systems. This meta benchmark combines seven commonly used short-answer grading datasets~\cite{dataset_stita, dataset_beetleII, dataset_saf, dataset_mohler, dataset_scientsbank, dataset_cunlp} containing questions, reference answers, provided (student) answers, and human grades normalized to the same grading scale.
For ASAG2024, we present initial evaluations of existing automated grading solutions on the benchmark. We show that \textit{specialized grading systems are still limited in their ability to generalize} to new questions and may need to be fine-tuned for specific use cases: \textit{their error is larger than a simple mean predictor baseline}. \textit{LLMs are able to generalize to the grading task} with decent performance without being specifically trained or fine-tuned to any specific grading data: \textit{mean error 0.27}. Aligned with related research, as the size of an LLM is increased, its ability to generalize to the grading task improves~\cite{emergent_abilities_of_llms}. 

\section{The ASAG2024 Benchmark}

The benchmark consists of seven SAG datasets in English and contains ~19'000 question-answer-grade triplets (see Table~\ref{tab:benchmark-comparison}). We scale the grades to lie between $0$ and $1$ to make results on the datasets comparable. Each dataset must at least contain reference answers, provided answers by humans and grades \cite{dataset_stita, dataset_beetleII, dataset_saf, dataset_mohler, dataset_scientsbank, dataset_cunlp}.

\begin{table}[ht]
  \caption{Datasets included in ASAG2024} 
  \label{table:overview_Quantitative}
  \small
  \centering
  \resizebox{1.0\hsize}{!}{\begin{tabular}{l|lllrlr}
    \toprule
    \textbf{Dataset} & 
    \textbf{Year} & 
    \textbf{Domain} & 
    \textbf{Ed. Level} & 
    \textbf{\# Entries} &
    \textbf{Grading Scale} & 
    \textbf{Mean Grade (scaled)} \\
    \midrule
    Beetle~\cite{dataset_beetleII} & 
    2014 & 
    Physics & 
    Upper secondary & 
    3941 &
    4 categories & 
    $0.67\pm 0.33$ \\
    \midrule
    CU-NLP~\cite{dataset_cunlp} & 
    2021 & 
    NLP & 
    Undergraduate & 
    171 &
    0-100 & 
    $0.28\pm 0.24$ \\
    \midrule
    DigiKlausur \cite{dataset_digiklausur} & 
    2019 & 
    Machine Learning & 
    Graduate & 
    646 &
    0-2 & 
    $0.68\pm 0.36$ \\
    \midrule
    Mohler \cite{dataset_mohler} & 
    2011 & 
    Data Structures & 
    Undergraduate & 
    630 &
    0-5 & 
    $0.81\pm 0.24$ \\
    \midrule
    SAF (English) \cite{dataset_saf} & 
    2022 & 
    Computer Science & 
    Undergraduate & 
    2463 &
    0-1 & 
    $0.76\pm 0.31$ \\
        \midrule
    SciEntsBank \cite{dataset_scientsbank} & 
    2012 & 
    Science Education & 
    Various & 
    10,804 &
    4 categories & 
    $0.60\pm 0.41$ \\

    \midrule
    Stita \cite{dataset_stita} & 
    2022 & 
    Statistics & 
    Undergraduate & 
    333 &
    0-1 & 
    $0.68\pm 0.28$ \\

    \bottomrule
  \end{tabular}}
\label{tab:benchmark-comparison}
\rmspace
\end{table}

\section{Experimental Evaluation}

For our newly created ASAG2024 dataset, we implement and evaluate a set of seven automated grading methods.

\textbf{Methods.} We select two specialized grading systems, \path{BART-SAF}~\cite{lewis2019bart, BART-SAF} \& \path{PrometheusII-7B}~\cite{kim2024prometheus}, that are employing task-finetuned language models.
Additionally, we evaluate three size categories of LLMs to validate whether these models can generalize from their pretraining task to grading through in-context learning (ICL): \path{Llama-3-8B}, \path{GPT-3.5- turbo-0125} and \path{GPT-4o-2024-05-13}.
We use a simple prompt with an instruction, question, reference answer, and student's answer.
We also implement two baselines: (1) Nomic-embed-text-v1~\cite{nussbaum2024nomic}, a recent embedding model (using cosine-similarity); (2) A mean baseline 
that simply predicts the mean grade.

\textbf{Metrics.}
Root Mean Square Error (RMSE) is a common metric for 
reporting ASAG results. 
It measures the average magnitude of prediction errors.
Due to grade imbalance, systems that assign higher grades generally have smaller errors (e.g., mean predictor). To counteract this, we introduce a weighted RMSE (wRMSE) in which the grades are weighted by how often similar grades appear in the data source. Specifically, a weight is given to each entry according to the number of other entries within a 0.1 range. Each dataset is divided into ten ranges of 0.1, and all ranges receive an equal 10\% share of the total weight. If any range does not contain any entries, its weight is distributed equally to the other ranges.

\begin{equation}
\text{wRMSE} = \sqrt{\sum_{i=0}^{N} w_i \cdot (y_i - \hat{y}_i)^2}
\end{equation}

\begin{itemize}
\item $N$ is the number of observations in an individual dataset.
\item $y_i$ is the actual observation, in our case, the human grade.
\item $\hat{y}_i$ is the prediction, i.e. the predicted grade.
\item $w_i$ is the weight of an individual observation.
\end{itemize}

\subsection{Initial Results}

Table~\ref{tab:results_rmsd} shows the wRMSE of all methods. \path{GPT-3.5-turbo} and \path{GPT-4o}  outperform other approaches, even without more advanced prompting techniques. Surprisingly, the fine-tuned models (\path{BART-SAF}, \path{PrometheusII-7B}) perform worse (0.48 and 0.43) even than the simple mean predictor baseline (0.40). The purely embedding-based model (\path{Nomic-embed-text}) that uses cosine-similarity between the student and the reference answer performs even slightly better than smaller task-independent LLMs such as \path{Llama3-8b}.

\begin{table}
\caption{Comparison of various models across different data sources according to their wRMSE}
\label{tab:results_rmsd}
\centering
\small
\resizebox{1.0\hsize}{!}{
\begin{tabular}{@{}lccccccc@{}}
\toprule
\textbf{Dataset} & \textbf{Baseline} & \textbf{Nomic-embed-text} & \textbf{BART-SAF} & \textbf{PrometheusII-7B} & \textbf{Llama3-8B} & \textbf{GPT-3.5-turbo} & \textbf{GPT-4o}  \\
\midrule
Beetle       & 0.41 & 0.32 & 0.51 & 0.55 & 0.37 & 0.33 & \textbf{0.32} \\
CU-NLP       & 0.40 & \textbf{0.28} & 0.48 & 0.42 & 0.43 & 0.31 & \textbf{0.34} \\
DigiKlausur  & 0.45 & 0.39 & 0.53 & 0.40 & 0.42 & 0.33 & \textbf{0.27} \\
Mohler       & 0.43 & 0.24 & 0.44 & 0.36 & 0.27 & 0.25 & \textbf{0.22} \\
SAF          & 0.41 & 0.41 & 0.47 & 0.39 & 0.46 & 0.28 & \textbf{0.24} \\
SciEntsBank  & 0.39 & 0.37 & 0.50 & 0.51 & 0.38 & 0.33 & \textbf{0.31} \\
Stita        & 0.34 & 0.40 & 0.44 & 0.40 & 0.38 & 0.28 & \textbf{0.22} \\
\midrule
\textbf{Mean}         & 0.40 & 0.34 & 0.48 & 0.43 & 0.39 & 0.30 & \textbf{0.27} \\
\bottomrule
\end{tabular}}
\rmspace
\end{table}

\section{Conclusion and Future Work}

Grading systems are not mature enough yet to be used in a fully automated exam setting. The best system (GPT-4o) still exhibits an error more than double that of a human ($0.1$ according to SAF~\cite{dataset_saf}).
However, LLM-based methods show stable performance across datasets, making them applicable for self-study or as a support tool for grading.
In the future, we will expand the benchmark with more diverse question-answer datasets, specifically focusing on multilingual aspects. 
We also plan to provide a more thorough evaluation of automated grading solutions on our dataset, including investigating common ICL strategies (e.g., few-shots, chain of thought). 

\begin{acks}
This work was supported by OpenAI’s Researcher Access Program.
\end{acks}

\bibliographystyle{ACM-Reference-Format}
\bibliography{references}

\end{document}